\documentclass[sigconf,nonacm]{acmart}

\usepackage{amsmath,amsfonts}
\usepackage{algorithmic}
\usepackage{graphicx}
\usepackage{textcomp}
\usepackage{xcolor}
\usepackage{lmodern}
\usepackage{booktabs}
\usepackage{multirow}

\def\BibTeX{{\rm B\kern-.05em{\sc i\kern-.025em b}\kern-.08em
    T\kern-.1667em\lower.7ex\hbox{E}\kern-.125emX}}

\begin{document}

\title{EAGLE: Edge-Aware Graph Learning for Proactive Delivery Delay Prediction in Smart Logistics Networks}

\author{Zhiming Xue*}
\affiliation{%
  \institution{College of Engineering, Northeastern University}
  \city{Boston}
  \country{USA}
}
\email{xue.zh@northeastern.edu}

\author{Menghao Huo}
\affiliation{%
  \institution{School of Engineering, Santa Clara University}
  \city{Santa Clara}
  \country{USA}
}
\email{menghao.huo@alumni.scu.edu}

\author{Yujue Wang}
\affiliation{%
  \institution{Department of Chemistry and Chemical Biology, University of New Mexico}
  \city{Albuquerque}
  \country{USA}
}
\email{yujue@unm.edu}

\begin{abstract}
Modern logistics networks generate rich operational data streams at every warehouse node and transportation lane---from order timestamps and routing records to shipping manifests---yet predicting delivery delays remains predominantly reactive. Existing predictive approaches typically treat this problem either as a tabular classification task, ignoring network topology, or as a time-series anomaly detection task, overlooking the spatial dependencies of the supply chain graph. To bridge this gap, we propose a hybrid deep learning framework for proactive supply chain risk management. The proposed method jointly models temporal order-flow dynamics via a lightweight Transformer patch encoder and inter-hub relational dependencies through an Edge-Aware Graph Attention Network (E-GAT), optimized via a multi-task learning objective. Evaluated on the real-world DataCo Smart Supply Chain dataset, our framework achieves consistent improvements over baseline methods, yielding an F1-score of 0.8762 and an AUC-ROC of 0.9773. Across four independent random seeds, the framework exhibits a cross-seed F1 standard deviation of only 0.0089---a 3.8$\times$ improvement over the best ablated variant---achieving the strongest balance of predictive accuracy and training stability among all evaluated models.
\end{abstract}



\maketitle

\section{Introduction}
The proliferation of digital order management and IoT-enabled tracking systems in modern logistics networks has transformed supply chains into highly observable cyber-physical systems, generating multivariate time-series data at every warehouse node and transportation edge. However, operational responses to delivery delays remain largely reactive. Given that delayed shipments incur significant costs and cascading disruptions, proactive delay forecasting is a critical imperative \cite{r1}.

Recent predictive methodologies exhibit notable limitations. One track focuses on tabular machine learning (e.g., XGBoost) \cite{r1}, discarding spatial dependencies between geographic hubs. A second track employs spatiotemporal GNNs \cite{r2, r3, r4, r1}, but often lacks explicit modeling of seasonal order-flow dynamics. Furthermore, existing models frequently suffer from training instability, exhibiting high variance across random initializations. In real-world logistics, predictive consistency is as crucial as peak accuracy.

To address these limitations, we introduce a hybrid temporal-graph learning framework. We formulate delay prediction as a graph-based classification and regression problem, constructing a supply chain graph of geographic regions and order flows. The framework integrates a PatchTST-Lite temporal encoder \cite{r5} to capture seasonality and an Edge-Aware Graph Attention Network (E-GAT) to model spatial dependencies. By decoupling temporal and structural modeling and optimizing via a multi-task objective, our framework emphasizes training stability alongside predictive performance.

The main contributions of this paper are threefold:
\begin{itemize}
    \item A hybrid temporal-graph framework for delay prediction that effectively integrates patch-based temporal encoding with edge-aware spatial attention.
    \item A multi-task learning design that combines classification and regression objectives, providing beneficial regularization and improving overall model robustness.
    \item Extensive experiments on the DataCo Supply Chain dataset demonstrating that EAGLE achieves F1\,=\,0.8762 and AUC\,=\,0.9773 with cross-seed F1 std of only 0.0089---a 3.8${\times}$ reduction versus the strongest ablated variant (A2, std\,=\,0.0338)---confirming that decoupled temporal-structural modeling achieves the best trade-off between predictive accuracy and cross-seed consistency among graph-based approaches.
\end{itemize}

\section{Related Work}
Spatiotemporal GNNs (e.g., D2STGNN \cite{r4}, PDFormer \cite{r6}) establish strong baselines for traffic forecasting, while recent works \cite{r2, r3} extend these to logistics routing. However, they target physical road topologies rather than abstract order-fulfillment graphs. In IoT anomaly detection, GDN \cite{r7} and TranAD \cite{r8} excel on sensor benchmarks but target concurrent deviations rather than future SLA violations. On the supply chain side, Kosasih and Brintrup \cite{r9} demonstrate that GNNs can effectively uncover hidden supplier relationships and predict risk exposure in real-world supply chain networks, while Ahmed et al. \cite{r1} achieve effective delivery risk prediction on DataCo using tabular SOM-ANN. However, tabular approaches discard graph structure. Our framework bridges this gap by integrating temporal encoding, edge-aware attention, and multi-task learning.

\paragraph*{Training Stability under Class Imbalance}
SLA violation rates are typically below 10\%, causing standard models to degenerate. Prior work uses graph-based oversampling \cite{r10} or topology-aware margin losses \cite{r11}. We take a complementary approach: prior-informed bias initialisation anchors the starting probability, and the regression auxiliary task supplies dense gradient signal during sparse early training.

\section{Methodology}

\subsection{Problem Formulation and Graph Construction}
We define the logistics supply chain as a graph $\mathcal{G} = (\mathcal{V}, \mathcal{E})$, where each node $v \in \mathcal{V}$ represents a geographic region (e.g., a distribution hub or customer macro-region), and each edge $e_{uv} \in \mathcal{E}$ represents a shipping lane between regions $u$ and $v$. 

For each node $v$, we observe a multivariate feature sequence over a sliding window of $T$ time steps, denoted as $\mathbf{X}_v \in \mathbb{R}^{T \times d_{node}}$. The node features include order volume, mean scheduled transit time, transit time standard deviation, average discount rate, and previous delay days. Each edge $e_{uv}$ is associated with static or slowly-varying features $\mathbf{E}_{uv} \in \mathbb{R}^{d_{edge}}$, encompassing historical transit times and shipping mode distributions (e.g., air, ground, sea).

The objective is to predict, for each node $v$ at the current time step, a binary label $y_{class} \in \{0, 1\}$ indicating whether a delivery SLA will be violated, and a continuous value $y_{reg} \in \mathbb{R}^+$ representing the expected delay magnitude in days. The target label $y_{class}$ uses a \emph{relative delay formulation}: for each node $v$, we compute a per-node historical delay baseline $\mu_v$ from the training split. A node is labelled positive if its next-window average delay exceeds its own historical mean, i.e., $y_{class} = \mathbb{1}[\bar{d}_{next} > \mu_v]$. This design forces the model to detect \emph{anomalous} routing deterioration rather than memorizing which geographic hubs are structurally high-risk, yielding approximately equal proportions of persistently-negative and switching nodes (50\%/50\%) and a healthy positive label rate of ${\approx}6.2\%$. For zero-baseline nodes (no historical delay), the label reduces to the standard binary indicator ($d > 0$). Critically, to prevent temporal leakage, node features are always derived from the \emph{current} window $[t,\,t{+}14)$ while labels are derived from the strictly non-overlapping \emph{future} window $[t{+}14,\,t{+}28)$.

\subsection{EAGLE Architecture}

\begin{figure}[htbp]
  \centering
  \includegraphics[width=0.7\linewidth]{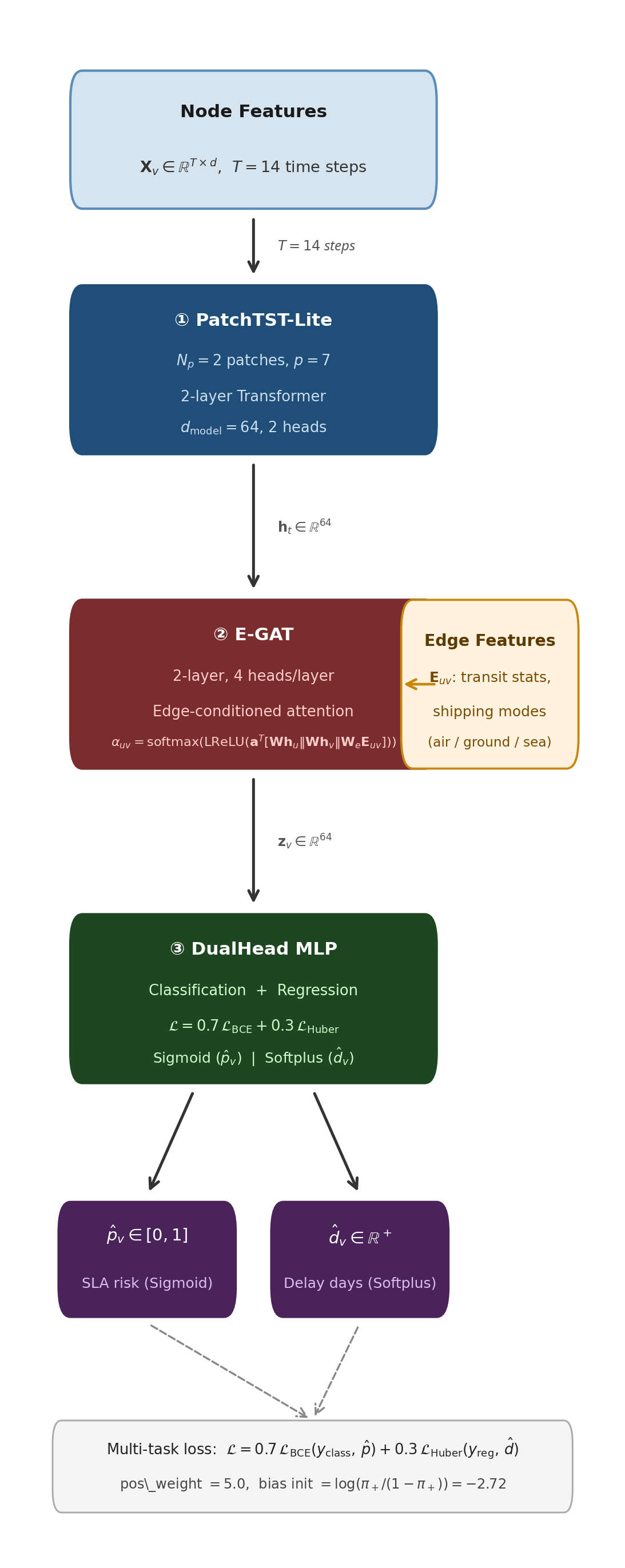}
  \caption{EAGLE three-module architecture pipeline.
  \textbf{(1) PatchTST-Lite:} Each node's $T{=}14$ time-step feature sequence
  is tokenised into $N_p{=}2$ non-overlapping patches of length $p{=}7$,
  encoded by a 2-layer Transformer (2 attention heads, $d_\text{model}{=}64$),
  and mean-pooled to a fixed temporal embedding $\mathbf{h}_t \in \mathbb{R}^{64}$.
  \textbf{(2) E-GAT:} A 2-layer Edge-Aware GAT (4 heads per layer) aggregates
  neighbourhood information, conditioning attention coefficients jointly on node
  embeddings \emph{and} static edge features $\mathbf{E}_{uv}$
  (historical transit statistics, shipping mode distributions).
  \textbf{(3) DualHead:} Two independent MLP heads produce the SLA-violation
  probability $\hat{p}_v \in [0,1]$ (sigmoid) and expected delay magnitude
  $\hat{d}_v \in \mathbb{R}^+$ (Softplus), trained end-to-end via a
  $0.7{\times}\text{BCE}+0.3{\times}\text{Huber}$ multi-task loss.}
  \Description{EAGLE three-module architecture pipeline.}
  \label{fig:arch}
\end{figure}

The EAGLE framework consists of three sequential modules: a temporal encoder, an edge-aware spatial encoder, and a dual-head prediction module.

\subsubsection{PatchTST-Lite Temporal Encoder}
To capture temporal dynamics without excessive overhead, we employ a channel-independent Transformer based on PatchTST \cite{r5}. For node $v$, the sequence $\mathbf{X}_v$ ($T{=}14$) is divided into non-overlapping patches ($p{=}7$), yielding $N_p{=}2$ tokens per channel. These are processed by a 2-layer Transformer ($d_{model}{=}64$, 2 heads) and mean-pooled to produce a temporal embedding $\mathbf{h}_v^{(t)} \in \mathbb{R}^{64}$. This approach, based on patches, accounts for bi-weekly seasonality and reduces the sensitivity to local noise, thus improving the robustness of temporal representations.

\subsubsection{Edge-Aware Graph Attention Network (E-GAT)}
The attention mechanism of Standard Graph Attention Networks (GATs) only considers node features. In supply chain networks, the features of the shipping lanes are important. Therefore, in this work we propose an Edge-Aware GAT (E-GAT) that incorporates edge features $\mathbf{E}_{uv}$. The attention coefficient $\alpha_{uv}$ is:
\begin{equation}
\alpha_{uv} = \text{softmax}_v \left( \text{LeakyReLU} \left( \mathbf{a}^T [\mathbf{W}\mathbf{h}_u^{(t)} \parallel \mathbf{W}\mathbf{h}_v^{(t)} \parallel \mathbf{W}_e \mathbf{E}_{uv}] \right) \right)
\end{equation}
where $\mathbf{W}, \mathbf{W}_e$ are weight matrices, $\mathbf{a}$ is the attention vector, and $\parallel$ is concatenation. The updated representation $\mathbf{z}_u$ is:
\begin{equation}
\mathbf{z}_u = \sigma \left( \sum_{v \in \mathcal{N}(u) \cup \{u\}} \alpha_{uv} \mathbf{W} \mathbf{h}_v^{(t)} \right)
\end{equation}
A two-layer E-GAT with 4 attention heads per layer is used to produce node embeddings \(\mathbf{Z} \in \mathbb{R}^{|\mathcal{V}|\times 64}\). 

\subsubsection{Dual-Head Prediction and Multi-Task Loss}
The node embedding \(\mathbf{z}_v\) is used as input to two independent MLPs. The first MLP corresponds to the classification task for the probability of SLA violation, \(\hat{p}_v = \text{Sigmoid}(\text{MLP}_{cls}(\mathbf{z}_v))\), and the second MLP corresponds to the regression task for delay magnitude, \(\hat{d}_v = \text{Softplus}(\text{MLP}_{reg}(\mathbf{z}_v))\). The combined loss is defined as:
\begin{equation}
\mathcal{L} = \lambda \, \mathcal{L}_{\mathrm{BCE}}\!\left(y_{\mathrm{class}},\, \hat{p}\right) {+} \left(1 - \lambda\right) \mathcal{L}_{\mathrm{Huber}}\!\left(y_{\mathrm{reg}},\, \hat{d}\right)
\end{equation}
We set $\lambda = 0.7$. This formulation acts as a strong regularizer; forcing shared representations to solve for both occurrence and magnitude prevents overfitting to binary labels, directly reducing performance variance.

\subsection{Explainability Module}
EAGLE provides dual-mode interpretability. \textbf{(1) Structural Risk Tracing:} E-GAT's attention weights $\alpha_{uv}$ are summed to create a risk heatmap. This helps operators identify the root cause hubs that propagate delay risk. \textbf{(2) Feature Attribution:} SHAP (KernelExplainer) is used to rank the contribution of input features to each prediction on the classification head.

\section{Experiments}

\subsection{Experimental Setup}
\textbf{Dataset:} We evaluate EAGLE on the publicly available DataCo Smart Supply Chain dataset \cite{r12}, comprising 180,519 order records with 53 features. We construct the supply chain graph by mapping shipping origins and customer regions to nodes. The resulting supply chain graph contains $N = 46$ geographic nodes and $E = 1478$ directed edges (739 undirected shipping lanes), representing active shipping lanes observed in the dataset. To ensure temporal validity and prevent data leakage, the dataset is split chronologically into training (70\%), validation (15\%), and testing (15\%) sets. The target label $y_{class}$ follows the relative delay formulation described in Section 3.1: a node is positive if its next-window average delay exceeds its own training-set baseline ($y_{class} = \mathbb{1}[\bar{d}_{next} > \mu_v]$), yielding a training-set positive rate of ${\approx}6.2\%$ (validation: 2.8\%, test: 4.0\%) and equal proportions of persistently-negative and occasionally-positive nodes.

\textbf{Baselines:} We compare EAGLE against four baselines forming a capability ladder:
\begin{itemize}
    \item \textit{Tabular Models:} XGBoost and Random Forest, representing the state-of-the-art for non-graph, order-level prediction on this dataset.
    \item \textit{Temporal Model:} LSTM, processing the temporal node features without graph structure.
    \item \textit{Graph Model:} Standard GAT, utilizing the static graph topology with mean-pooled temporal features (averaging each node's T=14 time steps into a single feature vector) as node input, but lacking explicit temporal encoding and edge features. This collapses all sequential information into a static representation, preventing the model from capturing temporal dynamics.
\end{itemize}
We omit GCN as GAT strictly generalizes it (node-only attention subsumes equal-weight aggregation), and GDN as it targets concurrent sensor anomaly detection rather than future discrete SLA violations.

\textbf{Implementation Details:} The proposed framework is implemented in PyTorch and PyTorch Geometric. The model is trained using the AdamW optimizer with a learning rate of 0.0003, Cosine Annealing learning rate schedule, and gradient clipping. To handle the approximately 6.2\% positive label rate, we apply a class-weighted BCE loss with pos\_weight\,$=$\,5.0 and initialise the classification head output bias to $\log(\pi_+/(1-\pi_+))\,{=}\,{-}2.72$, anchoring the starting prediction probability to the empirical label prior. We report the mean and standard deviation of metrics across 4 independent random seeds. All baseline methods are carefully tuned using comparable settings, including consistent training epochs, learning rates, and early stopping criteria to ensure fair comparison. To ensure fair comparison across model families, all methods are evaluated at the node-window level. For tabular models, order-level predictions are aggregated by averaging per-node probabilities within each label window. The same optimal-threshold calibration and relative label definition are applied to all methods.

\textbf{Data Pipeline Details.} Temporal snapshots are constructed using a sliding window of $T{=}14$ days with a stride of 1 day. For each window, five node-level features are aggregated from constituent orders: (1) order volume, (2) mean scheduled transit time, (3) standard deviation of scheduled transit time, (4) mean discount rate, and (5) mean actual delay days observed within the window. Edge features ($d_{edge}{=}7$) are computed once as static lane-level statistics: mean and standard deviation of scheduled transit time, order flow volume, and shipping mode distribution (one-hot fractions across four categories). A next-window protocol produces features from $[t, t{+}14)$ and labels from $[t{+}14, t{+}28)$, ensuring strict temporal separation.

After chronological splitting, the dataset comprises 698 training, 117 validation, and 191 test snapshots, each containing 46 nodes. The training set contains 1{,}975 positive and 30{,}133 negative node-window samples (positive rate 6.15\%); the validation set 153 positive / 5{,}229 negative (2.84\%); and the test set 351 positive / 8{,}435 negative (3.99\%). The lower positive rates in validation and test splits reflect natural temporal variation in delay patterns under chronological partitioning. For cold-start nodes with no historical delay ($\mu_v = 0$; 23 of 46 nodes), the relative label reduces to a simple binary delay indicator ($d > 0$), preserving label validity without requiring minimum order-volume thresholds.

\subsection{Performance Evaluation}
\begin{figure*}[htbp]
\centering
\includegraphics[width=0.9\textwidth]{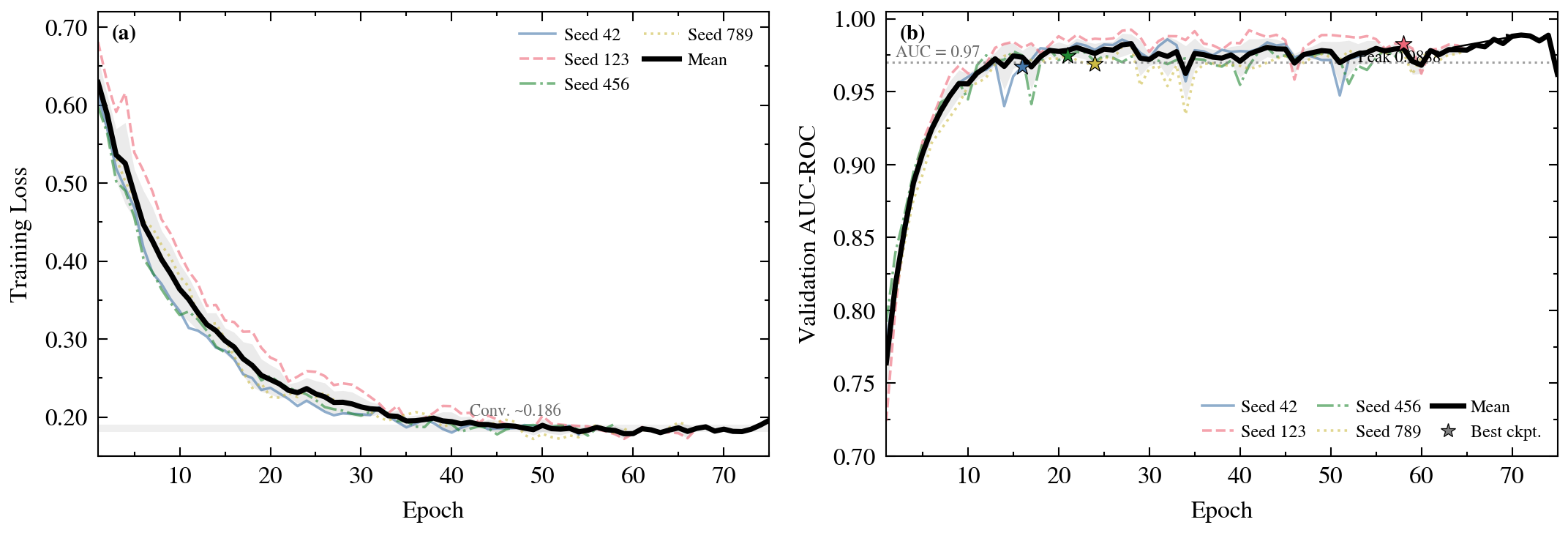}
\caption{EAGLE training dynamics across 4 random seeds. (a) Training loss converges smoothly to $\approx 0.186$ by epoch 40. (b) Validation AUC-ROC stabilises above 0.97 after 15 epochs. Stars mark the best-checkpoint epoch per seed.}
\Description{EAGLE training dynamics across 4 random seeds.}
\label{fig:training}
\end{figure*}

Comparative results on the DataCo test set are in Table \ref{tab:main_results}. Fig. \ref{fig:training} illustrates EAGLE's training dynamics across four seeds, demonstrating stable convergence.

\begin{table}[htbp]
\caption{Performance comparison on the DataCo test set (mean $\pm$  std, 4 seeds). All methods are evaluated at the node-window level with 14-day sliding window snapshots and an identical relative label definition. For tabular models (XGBoost, RF), predicted probabilities at the order level are averaged within each (node, label window) pair.}
\begin{center}
\setlength{\tabcolsep}{4pt}
\begin{tabular}{lcc}
\toprule
\textbf{Method} & \textbf{F1 (Macro)} & \textbf{AUC-ROC} \\
\midrule
XGBoost & $0.6379 \pm 0.0074$ & $0.7455 \pm 0.0020$ \\
Random Forest & $0.6052 \pm 0.0018$ & $0.7249 \pm 0.0032$ \\
LSTM & $0.8095 \pm 0.0035$ & $0.9679 \pm 0.0013$ \\
GAT & $0.6142 \pm 0.0789$ & $0.6872 \pm 0.0224$ \\
\midrule
\textbf{EAGLE (Ours)} & $\mathbf{0.8762 \pm 0.0089}$ & $\mathbf{0.9773 \pm 0.0037}$ \\
\bottomrule
\end{tabular}
\label{tab:main_results}
\end{center}
\end{table}

EAGLE outperforms all baselines under unified node-window evaluation. The results trace a clear capability hierarchy: moving from XGBoost to LSTM (+17.2 pp in F1) demonstrates the value of temporal sequence modeling; the standalone GAT lags LSTM by 19.5 pp, revealing that static graph topology without temporal context actually degrades performance; and moving from LSTM to EAGLE (+6.7 pp) isolates the incremental benefit of edge-aware graph attention, as both models share equivalent temporal encoding—confirming that spatial structure provides genuine complementary signal beyond what temporal modeling alone captures.

\paragraph*{Training Stability Analysis}
EAGLE achieves F1 std $= 0.0089$ across four seeds. We note that the purely temporal LSTM baseline exhibits even lower F1 std (0.0035), consistent with its simpler architecture and smaller parameter space. However, LSTM's lower variance comes at the cost of 8.2\% lower F1. EAGLE achieves the best accuracy-stability trade-off: its variance remains 3.8$\times$ lower than the strongest ablated graph variant (A2, std $= 0.0338$) while delivering the highest F1 overall. The standalone GAT exhibits F1 std $= 0.0789$, confirming graph attention without temporal grounding is highly sensitive to initialisation. Fig. \ref{fig:training} shows EAGLE's validation AUC-ROC stabilises above 0.97 after 15 epochs, with narrow inter-seed spread.

\subsection{Ablation Study}
To validate the contribution of each component in EAGLE, we conducted an ablation study (Table \ref{tab:ablation}).

\begin{table*}[htbp]
\caption{Ablation Study Results, and MAE (Days) for the regression sub-task.}
\begin{center}
\begin{tabular}{lccc}
\toprule
\textbf{Variant} & \textbf{F1 (Macro)} & \textbf{AUC-ROC} & \textbf{MAE (Days)} \\
\midrule
A1: No Temporal & $0.6851 \pm 0.0589$ & $0.8605 \pm 0.0698$ & $0.1801$ \\
A2: No Edge Features & $0.8027 \pm 0.0338$ & $0.9540 \pm 0.0462$ & $0.0269$ \\
A3: Single-Task & $0.8230 \pm 0.0341$ & $0.9738 \pm 0.0120$ & $0.0731^\dag$ \\
\midrule
\textbf{EAGLE (Full)} & $\mathbf{0.8762 \pm 0.0089}$ & $\mathbf{0.9773 \pm 0.0037}$ & $\mathbf{0.0185}$ \\
\bottomrule
\multicolumn{4}{l}{\footnotesize $^\dag$ A3 has no regression head; MAE equals the zero-prediction baseline.}
\end{tabular}
\label{tab:ablation}
\end{center}
\end{table*}

\textbf{A1 (No Temporal):} Replacing PatchTST-Lite with static features causes a drastic F1 drop (0.8762$\rightarrow$0.6851) and raises F1 std to 0.0589, underscoring temporal encoding as the primary driver of accuracy and stability. Notably, A1's regression MAE of 0.1801 exceeds the zero-prediction baseline (0.0731), indicating that without temporal context, the regression head produces erratic magnitude estimates—further evidence that temporal encoding is essential for both classification and regression sub-tasks. \textbf{A2 (No Edge Features):} Reverting to standard GAT reduces F1 to 0.8027; A2's F1 std of 0.0338 vs.\ EAGLE's 0.0089 shows a 3.8$\times$ variance reduction from edge-conditioned attention. \textbf{A3 (Single-Task):} The removal of the regression head leads to a decrease in F1 to 0.8230 and a rise in standard deviation to 0.0341, thus validating the positive effect of dense regression gradients. The regression head offers accurate delay predictions, as validated by the MAE of 0.0731 for A3’s predictions, which effectively predict a zero delay. EAGLE’s predictions attain a superior MAE = 0.0185. Among the three experiments, temporal encoding makes the largest individual contribution to the model’s accuracy, as measured by the change in F1 ($\Delta$F1\,=\,+0.191). Edge-aware attention and multi-task regularization contribute smaller but significant gains to the model’s accuracy, i.e., +0.074 and +0.053, respectively. These contribute to the 3.8$\times$--6.6$\times$ reduction in variance that differentiates EAGLE from its ablated versions.

\subsection{Interpretability Case Study}
EAGLE's attention weights enable interpretable risk diagnosis. Fig. 3 shows the risk graph for the supply chain, with nodes 10, 22, 17 exhibiting high centrality and risk—a finding that aligns with domain knowledge of critical routing hubs.

\begin{figure}[htbp]
  \centering
  \includegraphics[width=\linewidth]{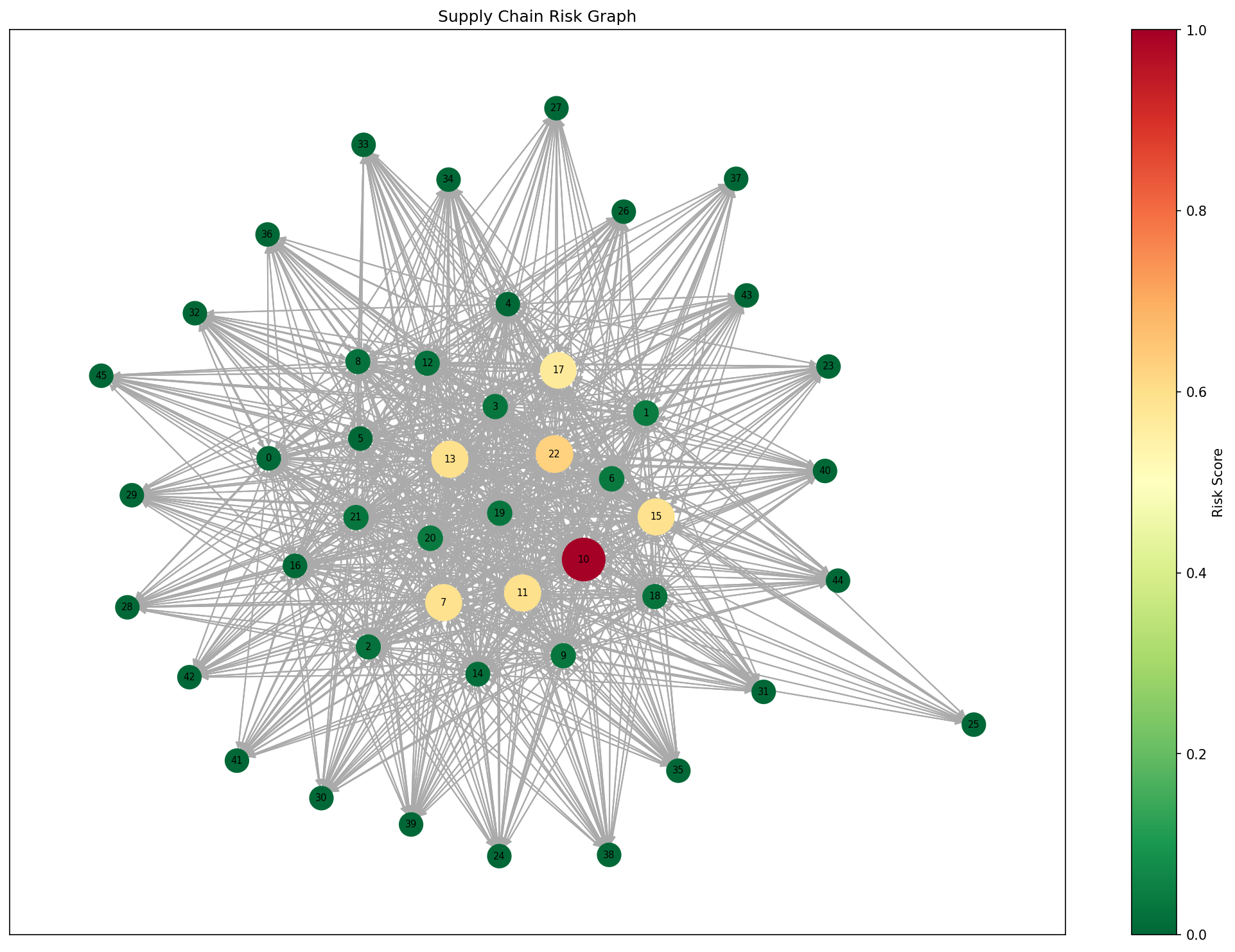}
  \caption{Supply chain risk graph with per-node risk scores aggregated across all test snapshots. Node color and size encode the normalized risk score (darker red = higher risk); edges represent active shipping lanes. High-risk hubs (Nodes 10, 22, 17) cluster at the network core, while peripheral low-volume nodes maintain consistently low risk.}
  \Description{Supply chain risk graph showing per-node normalized risk scores. Node 10 at the network core has the highest risk score of 1.00, shown in red.}
  \label{fig:risk_graph}
\end{figure}

\subsection{Data Integrity and Leakage Prevention}
In building a valid benchmark from DataCo, three types of leakage were eliminated. (i) \textbf{Direct label leakage}: Features such as \texttt{delivery\_status} and \texttt{days\_for\_shipping\_real} were direct encodings of the label and were removed. (ii) \textbf{Co-derivation leakage}: The feature \texttt{late\_delivery\_risk} is algebraically derived from scheduled days and results in feature/label correlation $r > 0.99$. Replacing this feature with actual delivery outcomes reduced all correlation values to $|r| < 0.50$. (iii) \textbf{Temporal leakage}: Overlapping windows were addressed using a "next window" protocol wherein features are defined over $[t, t{+}14)$ and labels are derived from the non-overlapping window $[t{+}14, t{+}28)$.

Table \ref{tab:feature_audit} provides a complete audit of every feature used by EAGLE, specifying its temporal scope and data source. The five node-level features are computed exclusively from the feature window $[t, t{+}14)$, which is strictly disjoint from the label window $[t{+}14, t{+}28)$. The per-node baseline $\mu_v$ is computed exclusively from the training split (70\% of data by chronological order), ensuring no future labels influence the relative-label definition. Edge features (scheduled transit statistics and shipping mode fractions) are computed as global aggregates across all orders. Note that while this technically includes orders from both validation and test periods, these features are actually descriptive of shipping lane attributes (scheduled transit time, mode of transport) and not results of delivery. These are known at the time of order placement and do not actually encode results of delivery. Thus, the argument for leakage prevention is more structural than statistical: outcome-encoding columns are completely absent, all temporal node features use strict past windows, and the node baseline is entirely based on training data.

\begin{table*}[t]
\caption{Feature audit. Temporal bounds and source for features are recorded to ensure that outcome information does not leak into the features.}
\begin{center}
\footnotesize
\setlength{\tabcolsep}{4pt}
\begin{tabular}{llllll}
\toprule
\textbf{Feature} & \textbf{Type} & \textbf{Source} & \textbf{Time Scope} & \textbf{Future Info?} & \textbf{Justification} \\
\midrule
order\_vol & Node & count(orders) & $[t, t+14)$ & No & Feature window only \\
mean\_scheduled\_transit & Node & mean(scheduled\_days) & $[t, t+14)$ & No & Known at order placement \\
std\_scheduled\_transit & Node & std(scheduled\_days) & $[t, t+14)$ & No & Known at order placement \\
mean\_discount\_rate & Node & mean(discount) & $[t, t+14)$ & No & Known at order placement \\
prev\_delay\_days & Node & mean(actual\_delay) & $[t, t+14)$ & No & Past-realised outcomes only \\
transit\_mean & Edge & mean(scheduled\_days) & Global static & No & Scheduled (not actual) transit; stable over time \\
transit\_std & Edge & std(scheduled\_days) & Global static & No & Same as above \\
flow\_volume & Edge & count(orders) & Global static & No & Lane-level aggregate, not outcome-related \\
mode\_distribution ($\times 4$) & Edge & fraction(mode) & Global static & No & Shipping mode choice, not delivery outcome \\
$\mu_v$ (baseline) & Label & mean(delay\_days) & Train split only & No & Computed exclusively from training data \\
$y_{class}$ & Label & $\mathbb{I}[d > \mu_v]$ & $[t+14, t+28)$ & N/A & Target variable \\
$y_{reg}$ & Label & mean(delay\_days) & $[t+14, t+28)$ & N/A & Target variable \\
\bottomrule
\addlinespace[2pt]
\multicolumn{6}{l}{\scriptsize Edge features are computed as global aggregations over scheduled (not outcome) statistics across all orders; see Section 4.5.}
\end{tabular}
\label{tab:feature_audit}
\end{center}
\end{table*}

\section{Discussion and Conclusion}
Two principal findings emerge. First, the results trace a clear capability hierarchy: tabular (0.64) $\rightarrow$ temporal (0.81) $\rightarrow$ temporal-graph (0.88), confirming that both temporal context and edge-aware spatial aggregation are necessary for effective logistics modeling. Second, EAGLE achieves the most favourable accuracy-stability trade-off among all models: it delivers the highest F1 (0.8762) while maintaining variance comparable to the simpler LSTM baseline (std 0.0089 vs.\ 0.0035). Among graph-aware models specifically, EAGLE's variance is 3.8$\times$ lower than the best ablated variant (A2) and 6.6$\times$ lower than the no-temporal variant (A1), confirming that temporal encoding, edge-aware attention, and multi-task regularization jointly suppress instability.

\textbf{Limitations.} Model complexity is higher than classical baselines, which may require more computational resources for large-scale deployment. Additionally, the relative-label formulation requires sufficient historical order volume per node to compute a stable per-node baseline $\mu_v$; for newly-established shipping lanes or cold-start nodes with sparse history, the label degenerates to a simple binary delay indicator, potentially reducing sensitivity to early routing deterioration. Generalization to logistics datasets with substantially different topological structures or order-volume distributions requires further validation.

\textbf{Conclusion.} This paper presents a hybrid temporal-graph learning framework for early delivery delay warning in IoT-enabled supply chains. Experimental results demonstrate consistent performance (F1=0.8762, AUC=0.9773), outperforming tabular and standard deep learning baselines by a notable margin while exhibiting improved training stability. Future work will explore scalability to larger supply chain graphs and real-time streaming inference.

\end{document}